\documentclass[10pt,twocolumn,letterpaper]{article}

\usepackage{cvpr}
\usepackage{times}
\usepackage{epsfig}
\usepackage{graphicx}
\usepackage{amsmath}
\usepackage{amssymb}
\usepackage{multirow}
\usepackage{booktabs}
\usepackage{tabu}
\usepackage{spreadtab}
\usepackage{tabularx}

\DeclareMathOperator*{\argmin}{arg\,min}

\usepackage[pagebackref=true,breaklinks=true,letterpaper=true,colorlinks,bookmarks=false]{hyperref}

\cvprfinalcopy 

\def\cvprPaperID{7904} 

\ifcvprfinal\fi
\begin{document}

\title{KeypointNet: A Large-scale 3D Keypoint Dataset\\ Aggregated from Numerous Human Annotations}

\author{Yang You, Yujing Lou\thanks{These authors contributed equally.}\ , Chengkun Li\footnotemark[1]\ , Zhoujun Cheng, Liangwei Li,\\ Lizhuang Ma, Weiming Wang\thanks{Weiming Wang is the corresponding author.}\ , Cewu Lu \\ 
Shanghai Jiao Tong University, China\\ 
}

\maketitle

\begin{abstract}
    Detecting 3D objects keypoints is of great interest to the areas of both graphics and computer vision. There have been several 2D and 3D keypoint datasets aiming to address this problem in a data-driven way. These datasets, however, either lack scalability or bring ambiguity to the definition of keypoints. Therefore, we present \textbf{KeypointNet}: the first large-scale and diverse 3D keypoint dataset that contains 103,450 keypoints and  8,234 3D models from 16 object categories, by leveraging numerous human annotations.
    To handle the inconsistency between annotations from different people, we propose a novel method to aggregate these keypoints automatically, through minimization of a fidelity loss. 
     Finally, ten state-of-the-art methods are benchmarked on our proposed dataset. Our code and data are available on \href{https://github.com/qq456cvb/KeypointNet}{https://github.com/qq456cvb/KeypointNet}.
\end{abstract}

\section{Introduction}

Detection of 3D keypoints is essential in many applications such as object matching, object tracking, shape retrieval and registration~\cite{mian2006three, bueno2016detection, wang2018learning}. Utilization of keypoints to match 3D objects has its advantage of providing features that are semantically significant and such keypoints are usually made invariant to rotations, scales and other transformations.
\begin{figure}[t]
    \centering
    \includegraphics[width=0.9\linewidth]{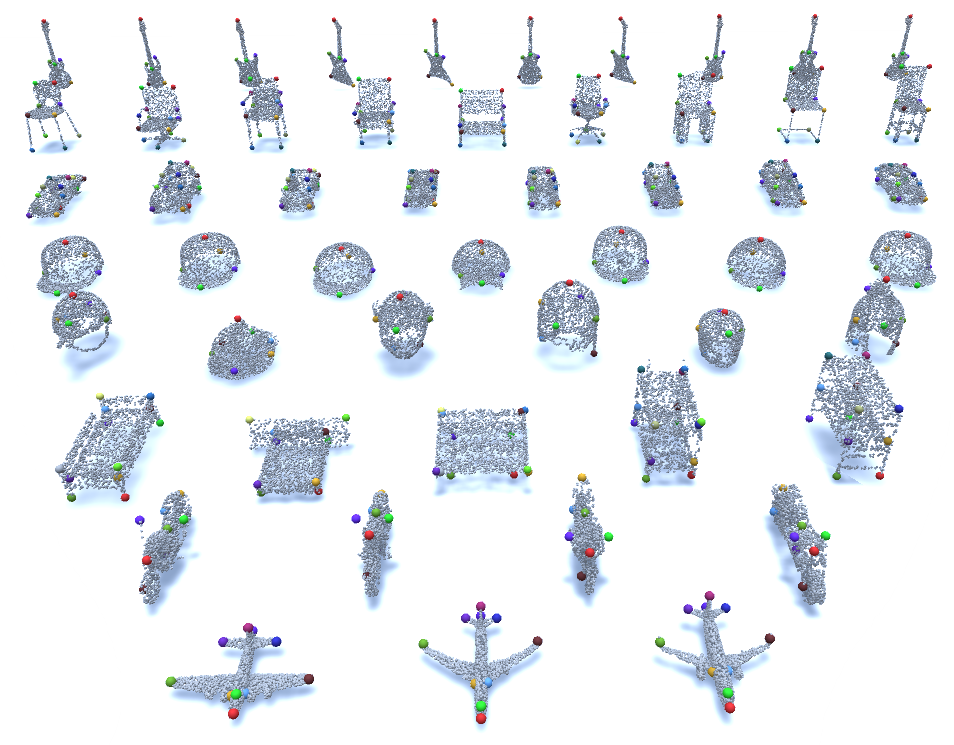}
    \caption{\textbf{We propose a large-scale KeypointNet dataset.} It contains 8K+ models and 103K+ keypoint annotations.}
    \label{fig:intro}
\end{figure}

In the trend of deep learning, 2D semantic point detection has been boosted with the help of a large quantity of high-quality datasets~\cite{andriluka14cvpr, min2019spair}. However, there are few 3D datasets focusing on the keypoint representation of an object. Dutagaci et al.~\cite{dutagaci2012evaluation} collect 43 models and label them according to annotations from various persons. Annotations from different persons are finally aggregated by geodesic clustering. ShapeNetCore keypoint dataset~\cite{yi2017syncspeccnn}, and a similar dataset~\cite{kim2013learning}, in another way, resort to an expert's annotation on keypoints, making them vulnerable and biased.

In order to alleviate the bias of experts' definitions on keypoints, we ask a large group of people to annotate various keypoints according to their own understanding. Challenges rise in that different people may annotate different keypoints and we need to identify the consensus and patterns in these annotations. 
Finding such patterns is not trivial when a large set of keypoints spread across the entire model. A simple clustering would require a predefined distance threshold and fail to identify closely spaced keypoints. As shown in Figure~\ref{fig:intro}, there are four closely spaced keypoints on each airplane empennage and it is extremely hard for simple clustering methods to distinguish them. Besides, clustering algorithms do not give semantic labels of keypoints since it is ambiguous to link clustered groups with each other. In addition, people's annotations are not always exact and errors of annotated keypoint locations are inevitable.
In order to solve these problems, we propose a novel method to aggregate a large number of keypoint annotations from distinct people, by optimizing a fidelity loss. After this auto aggregation process, we verify these generated keypoints based on some simple priors such as symmetry. 

In this paper, we build the first large-scale and diverse dataset named \textbf{KeypointNet} which contains 8,234 models with 103,450 keypoints. These keypoints are of high fidelity and rich in structural or semantic meanings. Some examples are given in Figure~\ref{fig:intro}. We hope this dataset could boost semantic understandings of common objects.

In addition, we propose two large-scale keypoint prediction tasks: keypoint saliency estimation and keypoint correspondence estimation. We benchmark ten state-of-the-art
algorithms with mIoU, mAP and PCK metrics. Results show that the detection and identification of keypoints remain a challenging task.

\begin{table*}[ht]
\small
\begin{minipage}{\textwidth}
\vspace{-0.15in}
\centering
\begin{tabular}{l|c|c|c|c|c|c|c}
\hline
\textbf{Dataset}  & Domain & Correspondence &Template-free & Instances & Categories & Keypoints & Format\\
\hline
\textbf{FAUST}~\cite{Bogo:CVPR:2014}  &human &  $\surd$& $\times$&100 & 1& 689K& mesh\\
\textbf{SyncSpecCNN}~\cite{yi2017syncspeccnn}&chair &  $\surd$& $\times$ &6243 & 1& $\sim$60K& point cloud\\
\textbf{Dutagaci et al.}~\cite{dutagaci2012evaluation} &general&  $\times$ & $\surd$ & 43& 16& $<$1K &mesh\\
\textbf{Kim et al.}~\cite{kim2012exploring}& general& $\surd$& $\times$& 404 & 4& $\sim$3K& mesh\\
\textbf{PASCAL 3D+}~\cite{xiang_wacv14} &general & $\times$& $\times$& 36292 & 12& 150K+&RGB w. 3D model\\
\hline
\textbf{Ours}&general& $\surd$& $\surd$& 8234 & 16& 103K+&point cloud \& mesh\\
\hline
\end{tabular}
\bigbreak
\caption{\textbf{Comparison of 3d keypoint datasets}. \textbf{Correspondence} indicates whether keypoints are indexed correspondingly. \textbf{Template-free} indicates whether it avoids hardcoded keypoint templates.}
\label{tab:comparison}
\end{minipage}
\end{table*}

In summary, we make the following contributions:
\begin{itemize}
    \item To the best of our knowledge, we provide the first large-scale dataset on 3D keypoints, both in number of categories and keypoints.
    
    \item We come up with a novel approach on aggregating people's annotations on keypoints, even if their annotations are independent from each other.
    
    \item We experiment with ten state-of-the-art benchmarks on our dataset, including point cloud, graph, voxel and local geometry based keypoint detection methods.
\end{itemize}

\section{Related Work}

\subsection{Detection of Keypoints}
Detection of 3D keypoints has been a very important task for 3D object understanding which can be used in many applications, such as object pose estimation, reconstruction, matching, segmentation, etc. Researchers have proposed various methods to produce interest points on objects to help further objects processing. Traditional methods like 3D Harris~\cite{sipiran2011harris}, HKS~\cite{sun2009concise}, Salient Points~\cite{castellani2008sparse}, Mesh Saliency~\cite{lee2005mesh}, Scale Dependent Corners~\cite{novatnack2007scale}, CGF~\cite{khoury2017learning}, SHOT~\cite{tombari2010unique}, etc, exploit local reference frames (LRF) to extract geometric features as local descriptors. 
However, these methods only consider the local geometric information without semantic knowledge, which forms a gap between detection algorithms and human understanding.

Recent deep learning methods like SyncSpecCNN~\cite{yi2017syncspeccnn}, deep functional dictionaries~\cite{sung2018deep} are proposed to detect keypoints. Unlike traditional ones, these methods do not handle rotations well. Though some recent methods like S2CNN~\cite{cohen2018spherical} and PRIN~\cite{you2018prin} try to fix this, deep learning methods still rely on ground-truth keypoint labels annotated
by human with expert verification.

\subsection{Keypoint Datasets}
Keypoint datasets have its origin in 2D images, where plenty of datasets on human skeletons and object interest points are proposed. For human skeletons, MPII human pose dataset~\cite{andriluka14cvpr}, MSCOCO keypoint challenge~\cite{mscoco} and PoseTrack~\cite{andriluka2018posetrack} annotate millions of keypoints on humans. For more general objects, SPair-71k~\cite{min2019spair} contains 70,958 image pairs with diverse variations in viewpoint and scale, with a number of corresponding keypoints on each image pair. PUB~\cite{WahCUB_200_2011} provides 15 part locations on 11,788 images from 200 bird categories and PASCAL~\cite{bourdev2009poselets} provides keypoint annotations for 20 object categories. HAKE~\cite{li2019hake} provides numerous annotations on human interactiveness keypoints. ADHA~\cite{pang2018deep} annotates key adverbs in videos, which is a sequence of 2D images.



Keypoint datasets on 3D objects, include Dutagaci et al.~\cite{dutagaci2012evaluation}, SyncSpecCNN~\cite{yi2017syncspeccnn} and Kim et al.~\cite{kim2013learning}. Dutagaci et al.~\cite{dutagaci2012evaluation} aggregates multiple annotations from different people with an ad-hoc method while the dataset is extremely small. Though SyncSpecCNN~\cite{yi2017syncspeccnn}, Pavlakos et al. \cite{pavlakos20176} and Kim et al.~\cite{kim2013learning} give a relatively large keypoint dataset, they rely on a manually designed template of keypoints, which is inevitably biased and flawed. GraspNet~\cite{fang2019graspnet} gives dense annotations on 3D object grasping. The differences between theirs and ours are illustrated in Table~\ref{tab:comparison}.

\section{KeypointNet: A Large-scale 3D Keypoint Dataset}

\subsection{Data Collection}
KeypointNet is built on ShapeNetCore~\cite{chang2015shapenet}. ShapeNetCore covers 55 common object categories with about 51,300 unique 3D models. 

 We filter out those models that deviate from the majority and keep at most 1000 instances for each category in order to provide a balanced dataset. In addition, a consistent canonical orientation is established (e.g., upright and front) for every category because of the incomplete alignment in ShapeNetCore.


We let annotators determine which points are important, and same keypoint indices should indicate same meanings for each annotator. Though annotators are free to give their own keypoints, three general principles should be obeyed: (1) each keypoint should describe an object’s semantic information shared across instances of the same object category, (2) keypoints of an object category should spread over the whole object and (3) different keypoints have distinct semantic meanings. After that, we utilize a heuristic method to aggregate these points, which will be discussed in Section~\ref{sec:aggr}.


Keypoints are annotated on meshes and these annotated meshes are then downsampled to 2,048 points. Our final dataset is a collection of point clouds, with keypoint indices.


\subsection{Annotation Tools}
\begin{figure}[h]
    \centering
    \includegraphics[width=0.9\linewidth]{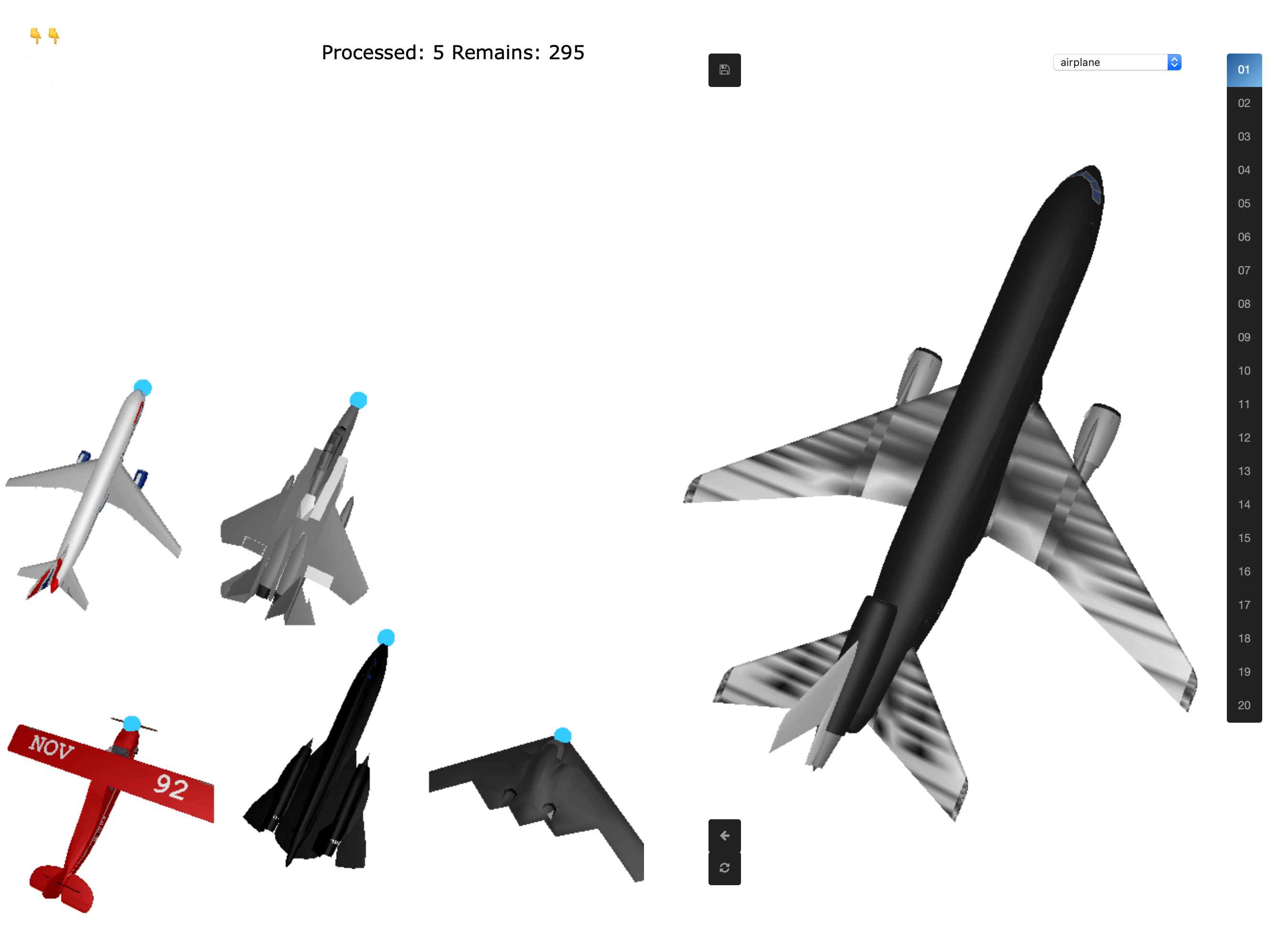}
    \caption{Web interface of the annotation tool.}
    \label{fig:web}
\end{figure}

We develop an easy-to-use web annotation tool based on NodeJS. Every user is allowed to click up to 24 interest points according to his/her own understanding. The UI interface is shown in Figure~\ref{fig:web}.  Annotated models are shown in the left panel while the next unprocessed model is shown in the right panel.

\begin{figure*}[ht]
    \centering
    \includegraphics[width=0.9\linewidth]{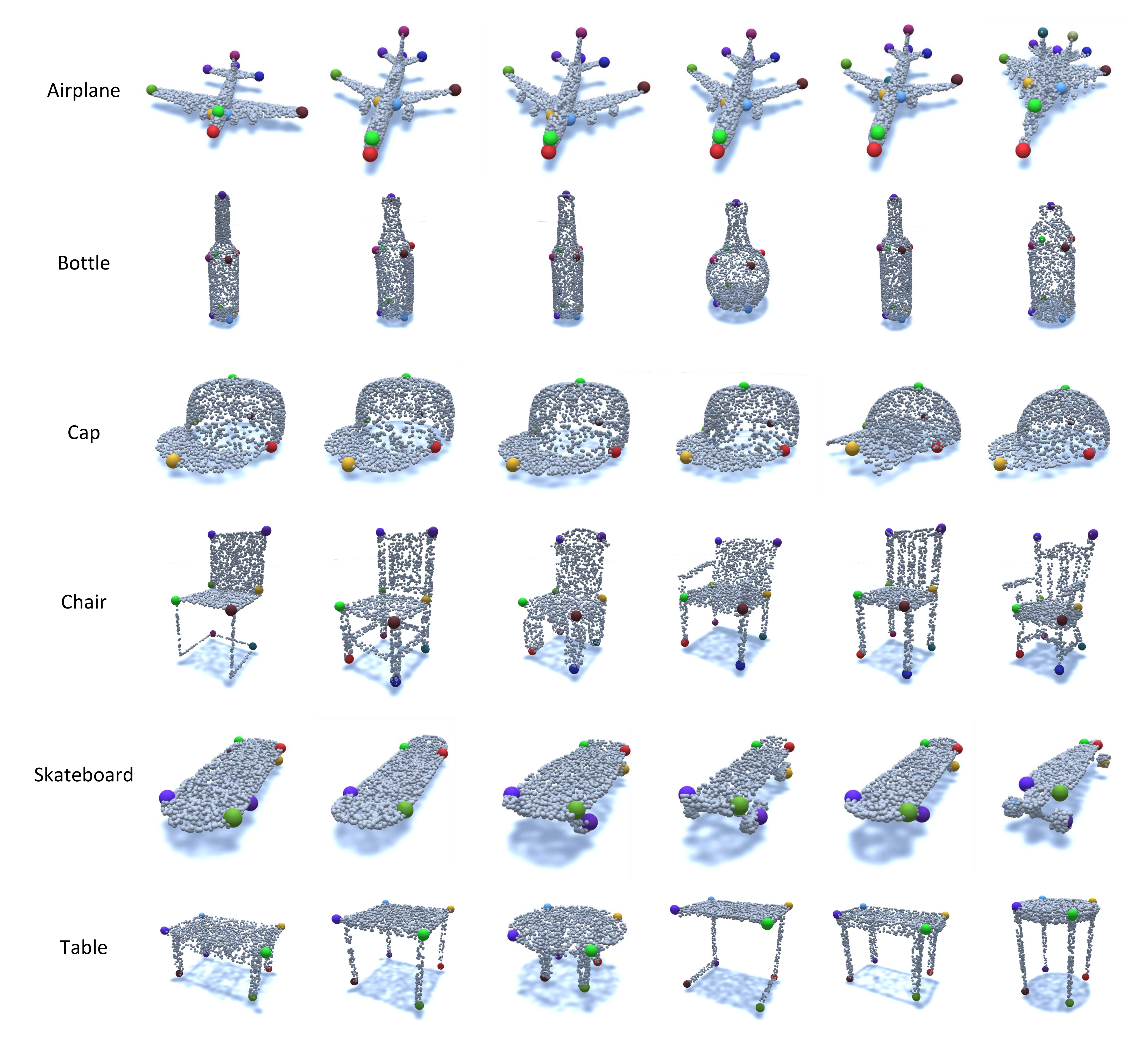}
    \caption{\textbf{Dataset Visualization.} Here we plot ground-truth keypoints for several categories. We can see that by utilizing our automatic aggregation method, keypoints of high fidelity are extracted.}
    \label{fig:vis}
\end{figure*}
 
\subsection{Dataset Statistics}
At the time of this work, our dataset has collected 16 common categories from ShapeNetCore, with 8234 models. Each model contains 3 to 24 keypoints. Our dataset is divided into train, validation and test splits, with 7:1:2 ratio.  Table~\ref{tab:statistics} gives detailed statistics of our dataset. Some visualizations of our dataset is given in Figure~\ref{fig:vis}.

\begin{figure*}[ht]
    \centering
    \includegraphics[width=0.9\linewidth]{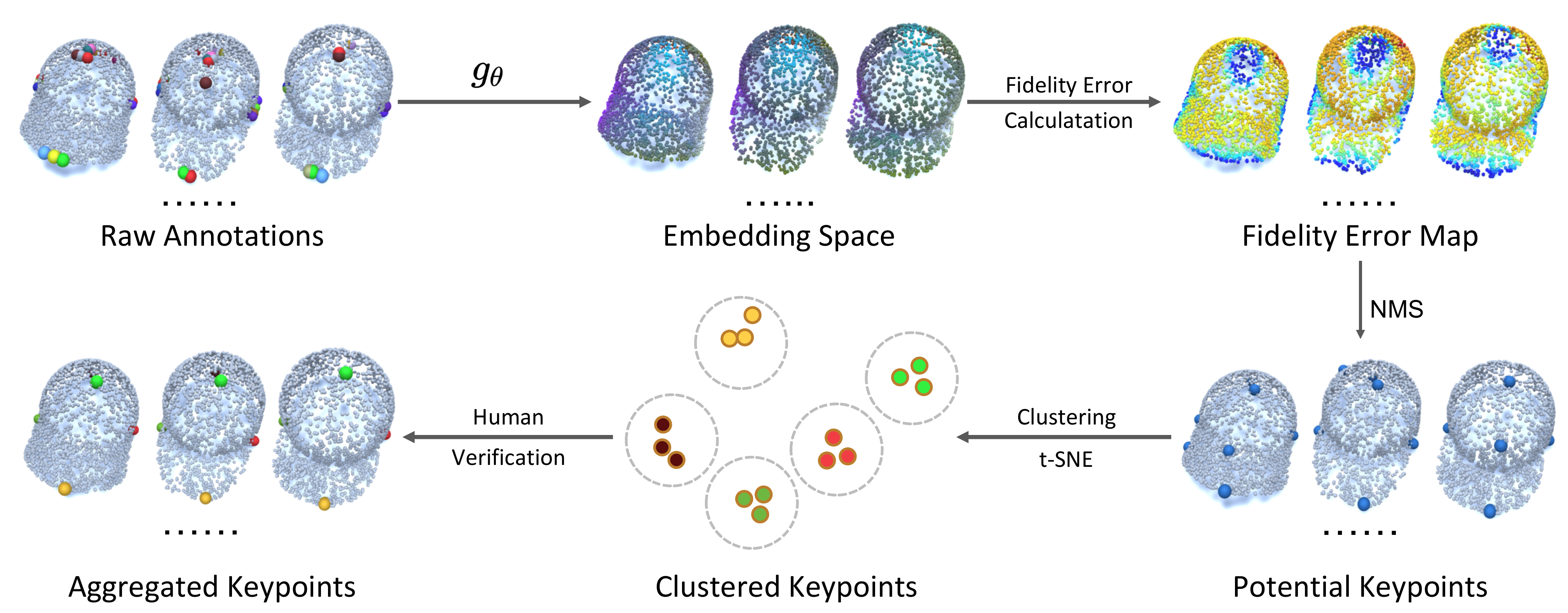}
    \caption{\textbf{Keypoint aggregation pipeline.} We first infer dense embeddings from human labeled raw annotations. Then fidelity error maps are calculated by summing embedding distances to human labeled keypoints. Non Minimum Suppression is conducted to form a potential set of keypoints. These keypoints are then projected onto 2D subspace with t-SNE and verified by humans.}
    \label{fig:pipeline}
\end{figure*}

\begin{table*}[ht]
\begin{minipage}[b]{.5\linewidth}
\begin{center}
\resizebox{!}{0.4\textwidth}{
\begin{spreadtab}{{tabular}{l|c|c|c|c|c}}
\hline
@\textbf{Category}  & @\textbf{Train} & @\textbf{Val} & @\textbf{Test} & @\textbf{All} & @\textbf{\#Annotators} \\
\hline
@\textbf{Airplane} & 715 & 102 & 205 & sum(b2:d2) & 26\\
@\textbf{Bathtub} & 344 & 49 & 99 & sum(b3:d3) & 15\\
@\textbf{Bed} & 102 & 14 & 30 & sum(b4:d4) & 8\\
@\textbf{Bottle} & 266 & 38 & 76 & sum(b5:d5) & 9\\
@\textbf{Cap} & 26 & 4 & 8 & sum(b6:d6) & 6\\
@\textbf{Car} & 701 & 100 & 201 & sum(b7:d7) & 17\\
@\textbf{Chair} & 699 & 100 & 200 & sum(b8:d8) & 15\\
@\textbf{Guitar} & 487 & 70 & 140 & sum(b9:d9) & 13\\
@\textbf{Helmet} & 62 & 10 & 18 & sum(b10:d10) & 8\\
@\textbf{Knife} & 189 & 27 & 54 & sum(b11:d11) & 5\\
@\textbf{Laptop} & 307 & 44 & 88 & sum(b12:d12) & 12\\
@\textbf{Motorcycle} & 208 & 30 & 60 & sum(b13:d13) & 7\\
@\textbf{Mug} & 130 & 18 & 38 & sum(b14:d14) & 9\\
@\textbf{Skateboard} & 98 & 14 & 29 & sum(b15:d15) & 9\\
@\textbf{Table} & 786 & 113 & 225 & sum(b16:d16) & 17\\
@\textbf{Vessel} & 637 & 91 & 182 & sum(b17:d17) & 18\\
\hline
@\textbf{Total} & sum(b2:b17) & sum(c2:c17) & sum(d2:d17) & sum(e2:e17) & sum(f2:f17)\\
\hline
\end{spreadtab}
}
\end{center}
\end{minipage}
\hfill
\begin{minipage}[b]{.5\linewidth}
\begin{center}
\resizebox{!}{0.4\textwidth}{
\begin{spreadtab}{{tabular}{l|c|c|c|c}}
\hline
@\textbf{Category}  & @\textbf{Train} & @\textbf{Val} & @\textbf{Test} & @\textbf{All} \\
\hline
@\textbf{Airplane} & 9695 & 1379 & 2756 & sum(b2:d2) \\
@\textbf{Bathtub} & 5519 & 772 & 1589 & sum(b3:d3) \\
@\textbf{Bed} & 1276 & 188 & 377 & sum(b4:d4) \\
@\textbf{Bottle} & 4366 & 625 & 1269 & sum(b5:d5) \\
@\textbf{Cap} & 154 & 24 & 48 & sum(b6:d6) \\
@\textbf{Car} & 14976 & 2133 & 4294 & sum(b7:d7) \\
@\textbf{Chair} & 8488 & 1180 & 2395 & sum(b8:d8) \\
@\textbf{Guitar} & 4112 & 591 & 1197 & sum(b9:d9) \\
@\textbf{Helmet} & 529 & 85 & 162 & sum(b10:d10) \\
@\textbf{Knife} & 946 & 136 & 276 & sum(b11:d11) \\
@\textbf{Laptop} & 1842 & 264 & 528 & sum(b12:d12) \\
@\textbf{Motorcycle} & 2690 & 394 & 794 & sum(b13:d13) \\
@\textbf{Mug} & 1427 & 198 & 418 & sum(b14:d14) \\
@\textbf{Skateboard} & 903 & 137 & 283 & sum(b15:d15) \\
@\textbf{Table} & 6325 & 913 & 1809 & sum(b16:d16) \\
@\textbf{Vessel} & 9168 & 1241 & 2579 & sum(b17:d17) \\
\hline
@\textbf{Total} & sum(b2:b17) & sum(c2:c17) & sum(d2:d17) & sum(e2:e17) \\
\hline
\end{spreadtab}
}
\end{center}
\end{minipage}
\bigbreak
\caption{\textbf{Keypoint Dataset statistics.} Left: number of models in each category. Right: number of keypoints in each category.}
\label{tab:statistics}
\end{table*}

\section{Keypoint Aggregation}
\label{sec:aggr}
Given all human labeled raw keypoints, we leverage a novel method to aggregate them together into a set of \textit{ground-truth} keypoints. 

There are generally two reasons: 1) distinct people may annotate different sets of keypoints and human labeled keypoints are sometimes erroneous, so we need an elegant way to aggregate these keypoints; 2) a simple clustering algorithm would fail to distinguish those closely spaced keypoints and cannot give consistent semantic labels.

\subsection{Problem Statement} 
Given a $2$-dimensional sub-manifold $\mathcal{M}_m\subset \mathbb{R}^{3}$, where $m$ is the index of the model, a valid annotation from the $c$-th person is a keypoint set $\{l_{m,k}^{(c)}|l_{m,k}^{(c)} \in\mathcal{M}_m\}_{k=1}^{K_c}$, where $k$ is the keypoint index and $K_c$ is the number of keypoints annotated by person $c$. Note that different people may have different sets of keypoint indices and these indices are independent.

Our goal is to aggregate a set of potential ground-truth keypoints $\mathcal{Y}=\{y_{m,k}|y_{m,k} \in \mathcal{M}_m, m=1,\dots M,k=1,\dots K_m \}$, where ${K_m}$ is the number of proposed keypoints for each model $\mathcal{M}_m$, so that $y_{m_1,k}$ and $y_{m_2,k}$ share the same semantic.

\subsection{Keypoint Saliency}
Each annotation is allowed to be erroneous within a small region, so that a keypoint distribution is defined as follows:
\begin{align*}
    p(x|x\text{ is the $k$-th keypoint}, x\in\mathcal{M}_m) = \frac{\phi(l_{m,k}, x)}{Z(\phi)},
\end{align*}
where $\phi$ is Gaussian kernel function. $Z$ is a normalization constant. 
This contradicts many previous methods on annotating keypoints where a $\delta$-function is implicitly assumed. We argue that it is common that humans make mistakes when annotating keypoints and due to central limit theorem, the keypoint distribution would form a Gaussian.


\subsection{Ground-truth Keypoint Generation}
We propose to jointly output a dense mapping function $g_\theta: \mathcal{M} \rightarrow \mathbb{R}^d$ whose parameters are $\theta$, and the aggregated ground-truth keypoint set $\mathcal{Y}$. $g_\theta$ transforms each point into an high-dimensional embedding vector in $\mathbb{R}^d$. Specifically, we solve the following optimization problem:
\begin{align}
\label{eq:main}
\begin{split}
    (\theta^\ast, \mathcal{Y}^\ast) &= \argmin_{\theta, \mathcal{Y}}[f(\mathcal{Y}, g_\theta)  + H(g_\theta)] \\
    &s.t.\ g_\theta(y_{m_1,k}) \equiv g_\theta(y_{m_2,k}), \forall m_1, m_2, k.
\end{split}
\end{align}
where $f(\mathcal{Y}, g_\theta)$ is the data fidelity loss and $H(g_\theta)$ is a regularization term to avoid trivial solution like $g_\theta \equiv 0$. The constraint states that the embedding of ground-truth keypoints with the same index should be the same.

\paragraph{Fidelity Loss}
We define $f(\mathcal{Y}, g_\theta)$ as:
\begin{align*}
    f(\mathcal{Y}, g_\theta) = \sum_{m=1}^M\sum_{c=1}^C\sum_{k=1}^{K_m}\int_{\mathcal{M}_m}\mathbf{d}_\theta(x, y_{m,k})\frac{\phi(l_{m,n^*}^{(c)}, x)}{Z(\phi)}dx,
\end{align*}
 where $\mathbf{d}_\theta$ is the L2 distance between two vectors in embedding space:
 \begin{align*}
     \mathbf{d}_\theta(a, b) = \|g_\theta(a) - g_\theta(b)\|_2^2,
 \end{align*}
and
\begin{align*}
    n^* = \argmin_{n} \mathbf{d}_\theta({l_{m,n}^{(c)}}, y_m).
\end{align*}
 
 Unlike previous methods such as Dutagaci et al.\cite{dutagaci2012evaluation} where a simple geodesic average of human labeled points is given as ground-truth points, we seek a point whose expected embedding distance to all human labeled points is smallest. The reason is that geodesic distance is sensitive to the misannotated keypoints and could not distinguish closely spaced keypoints, while embedding distance is more robust to noisy points as the embedding space encodes the semantic information of an object. 
 

Equation~\ref{eq:main} involves both $\theta$ and $\mathcal{Y}$ and it is impractical to solve this problem in closed form. In practice, we use alternating minimization with a deep neural network to approximate the embedding function $g_\theta$, so that we solve the following dual problem instead (by slightly loosening the constraints):

\begin{align}
\begin{split}
\label{eq:altth}
    \theta^\ast = &\argmin_\theta [H(g_\theta)
    \\&+ \lambda \sum_{m_1,m_2}^M\sum_k^{K_m}\|g_\theta(y_{m_1,k}) - g_\theta(y_{m_2,k})\|_2^2],
\end{split}
\end{align}

\begin{align}
\begin{split}
\label{eq:alty}
    \mathcal{Y}^\ast = \argmin_\mathcal{Y}& \sum_{m=1}^M\sum_{c=1}^C\sum_{k=1}^{K_m}\int_{\mathcal{M}_m}\mathbf{d}_\theta(x, y_{m,k})\frac{\phi(l_{m,n^\star}^{(c)}, x)}{Z(\phi)}dx,\\
    &s.t.\ g_\theta(y_{m_1,k}) \equiv g_\theta(y_{m_2,k}), \forall m_1, m_2, k,
\end{split}
\end{align}
and alternate between the two equations until convergence.

By solving this problem, we find both an optimal embedding function $g_\theta$, together with intra-class consistent ground-truth keypoints $\mathcal{Y}$, while keeping its embedding distance from human-labeled keypoints as close as possible. The ground-truth keypoints can be viewed as the projection of human labeled data onto embedding space.

\begin{figure*}[t]
    \centering
    \includegraphics[width=0.9\linewidth]{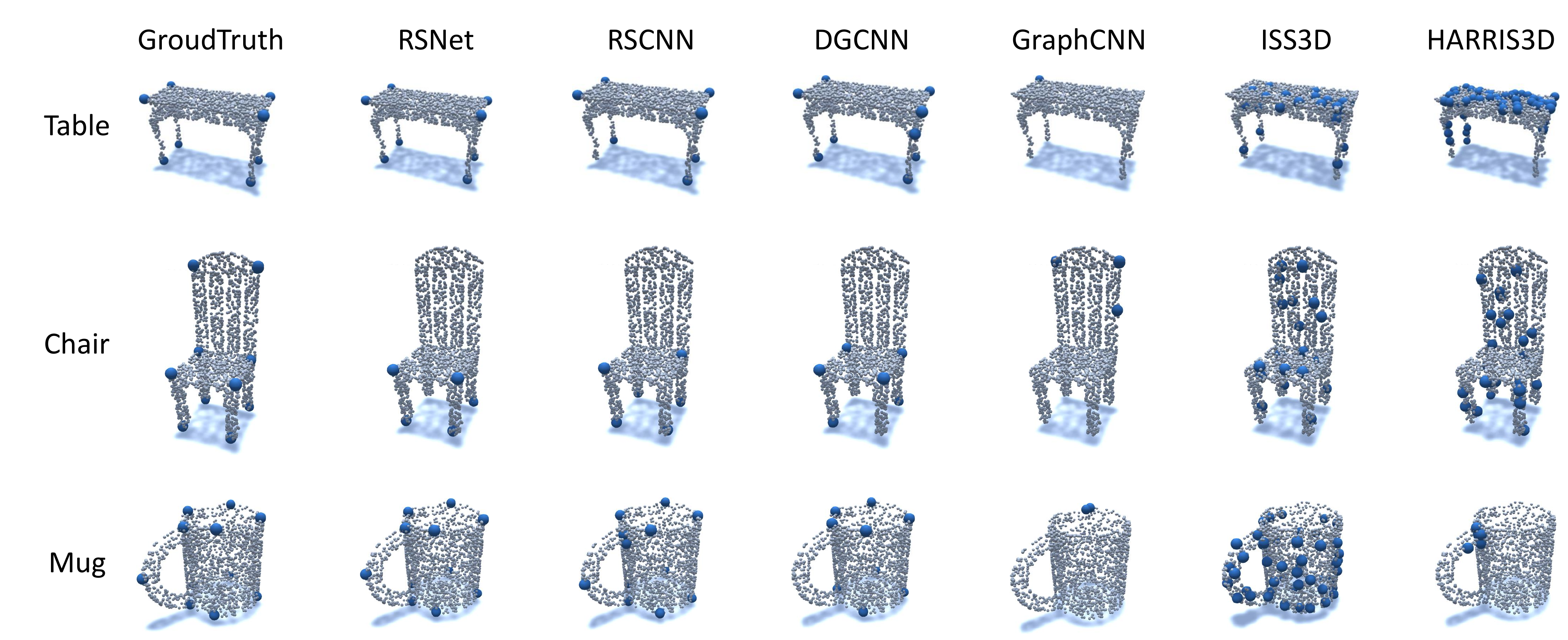}
    \caption{Visualizations of detected keypoints for six algorithms.}
    \label{fig:vis_detect}
\end{figure*}

\begin{table*}[t]
\begin{centering}
\resizebox{\textwidth}{!}{
\begin{tabular}{l|c|c|c|c|c|c|c|c|c|c|c|c|c|c|c|c|c}
\hline
~ & \textbf{Airplane} & \textbf{Bathtub} & \textbf{Bed} & \textbf{Bottle} & \textbf{Cap} & \textbf{Car} & \textbf{Chair} & \textbf{Guitar} & \textbf{Helmet} & \textbf{Knife} & \textbf{Laptop} & \textbf{Motor} & \textbf{Mug} & \textbf{Skate} & \textbf{Table} & \textbf{Vessel} & \textbf{Average}\\
\hline
\textbf{PointNet} & 10.4/9.3 & 0.8/6.3 & 3.0/9.4 & 0.5/8.2 & 0.0/0.3 & 7.3/9.2 & 7.7/6.5 & 0.1/1.4 & 0.0/1.3 & 0.0/0.9 & 23.6/32.0 & 0.1/4.0 & 0.2/3.9 & 0.0/2.9 & 20.9/23.1 & 7.1/9.1 & 5.1/8.0 \\
\textbf{PointNet++} & 33.5/44.6 & 18.5/32.3 & 16.4/23.4 & 22.1/38.2 & 13.2/12.1 & 24.0/41.0 & 18.2/29.4 & 24.3/27.7 & 7.0/7.4 & 19.0/22.1 & 30.9/60.0 & 21.9/36.0 & 16.3/21.4 & 13.7/16.1 & 29.3/47.4 & 16.7/26.9 & 20.3/30.4 \\
\textbf{RSNet} & 33.7/38.8 & 23.1/36.0 & 28.4/42.5 & 29.4/44.3 & \textbf{26.7}/\textbf{26.9} & 30.4/42.3 & 22.7/26.2 & 30.5/34.2 & 15.8/\textbf{21.3} & 21.7/31.5 & 39.8/62.0 & 31.5/43.2 & \textbf{24.1}/\textbf{35.2} & 21.6/25.9 & 39.7/54.6 & 18.3/22.3 & 27.3/36.7 \\
\textbf{SpiderCNN} & 24.6/24.7 & 13.0/16.4 & 16.4/17.1 & 3.6/8.2 & 0.0/0.5 & 11.8/17.2 & 18.2/20.4 & 7.4/9.9 & 0.0/0.5 & 14.9/18.7 & 25.6/40.0 & 12.0/13.3 & 2.2/4.0 & 1.8/3.3 & 27.9/40.5 & 12.9/14.6 & 12.0/15.6 \\
\textbf{PointConv} & \textbf{40.0}/51.2 & \textbf{29.0}/\textbf{44.7} & 25.1/\textbf{43.7} & 33.6/46.5 & 0.0/7.0 & \textbf{35.9}/\textbf{55.6} & \textbf{25.5}/\textbf{39.8} & 28.0/39.3 & 7.7/8.3 & 26.6/38.1 & 43.1/63.5 & \textbf{35.9}/\textbf{50.6} & 19.8/32.0 & 9.3/10.1 & \textbf{44.1}/\textbf{64.0} & 21.7/31.8 & 26.6/39.1 \\
\textbf{RSCNN} & 31.1/45.0 & 20.4/34.9 & 26.2/40.1 & 25.6/40.1 & 7.0/8.9 & 25.4/44.7 & 22.0/33.0 & 27.3/34.1 & 8.9/10.8 & 21.1/27.2 & 35.2/58.8 & 23.6/37.0 & 15.9/18.6 & 21.9/\textbf{30.1} & 27.4/49.8 & 18.1/28.9 & 22.3/33.9 \\
\textbf{DGCNN} & 39.2/\textbf{52.0} & 25.5/40.6 & \textbf{32.4}/42.8 & \textbf{40.3}/\textbf{61.6} & 0.0/2.0 & 27.0/48.6 & 24.9/34.6 & \textbf{31.9}/\textbf{44.1} & \textbf{17.1}/19.9 & \textbf{31.5}/\textbf{39.9} & \textbf{44.8}/\textbf{69.6} & 30.3/41.4 & 23.9/33.2 & \textbf{22.6}/29.8 & 41.0/59.0 & \textbf{24.4}/\textbf{35.6} & \textbf{28.5}/\textbf{40.9} \\
\textbf{GraphCNN} & 0.7/0.7 & 17.0/23.7 & 22.2/31.5 & 23.3/42.6 & 0.0/4.1 & 23.0/36.5 & 0.6/0.6 & 17.0/21.6 & 9.7/12.5 & 18.7/23.3 & 33.3/49.7 & 0.6/0.7 & 0.5/0.6 & 11.7/12.8 & 24.5/34.7 & 15.1/19.3 & 13.6/19.7 \\
\hline
\textbf{Harris3D} & 1.6/- & 0.3/- & 0.9/- & 0.5/- & 0.0/- & 0.9/- & 0.7/- & 1.2/- & 0.0/- & 1.9/- & 0.3/- & 0.7/- & 0.0/- & 1.0/- & 0.4/- & 1.2/- & 0.7/- \\
\textbf{SIFT3D} & 1.2/- & 1.0/- & 0.5/- & 1.1/- & 0.4/- & 1.0/- & 0.7/- & 0.7/- & 0.4/- & 0.6/- & 0.2/- & 0.8/- & 0.5/- & 0.7/- & 0.3/- & 1.0/- & 0.7/- \\
\textbf{ISS3D} & 1.1/- & 1.4/- & 0.9/- & 1.8/- & 0.6/- & 1.7/- & 0.7/- & 0.7/- & 0.8/- & 0.4/- & 0.1/- & 0.8/- & 0.9/- & 0.8/- & 0.3/- & 1.1/- & 0.9/- \\
\hline
\end{tabular}}
\end{centering}   
\caption{mIoU and mAP results (in percentage) for compared methods with distance threshold 0.01.}
\label{tab:map}
\end{table*}

\paragraph{Non Minimum Suppression}
Equation~\ref{eq:alty} may be hard to solve since $K_m$ is also unknown beforehand. For each model $\mathcal{M}_m$, the fidelity error associated with each potential keypoint $y_m\in \mathcal{M}_m$ is:
\begin{align}
    f(y_m, g_\theta) = \sum_{m'\neq m}^M\sum_{c=1}^C\int_{\mathcal{M}_{m'}}\mathbf{d}_\theta(x, y_{m'})\frac{\phi(l_{{m'},n^*}^{(c)}, x)}{Z(\phi)}dx,
\end{align}
where $y_{m'} = \argmin_{y_{m'}\in\mathcal{M}_{m'}}\|g_\theta(y_{m'}) - g_\theta(y_{m})\|_2^2$.

Then $y_m^\ast$ is found by conducting Non Minimum Suppression (NMS), such that:

\begin{align}
\begin{split}
     f(y_m^\ast, g_\theta) \leq  f(y_m, g_\theta), \\\forall y_m \in \mathcal{M}_m, \mathbf{d}_\theta(y_m, y_m^{*}) < \delta,
\end{split}
\end{align}
where $\delta$ is some neighborhood threshold.



After NMS, we would get several ground-truth points $y_{m,1}, y_{m,2}, \dots, y_{m,k}$ for each manifold $\mathcal{M}_m$. However, the arbitrarily assigned index $k$ within each model does not provide a consistent semantic correspondence across different models. Therefore we cluster these points according to their embeddings by first projecting them onto 2D subspace with t-SNE~\cite{maaten2008visualizing}.

\paragraph{Ground-truth Verification}
Though the above method automatically aggregate a set of potential set of keypoints with high precision, it omits some keypoints in some cases. As the last step, experts manually verify these keypoints based on some simple priors such as the rotational symmetry and centrosymmetry of an object.

\subsection{Implementation Details}
At the start of the alternating minimization, we initialize $\mathcal{Y}$ to be sampled from raw annotations and then run one iteration, which is enough for the convergence. We choose PointConv with hidden dimension 128 as the embedding function $g$. During the optimization of Equation~\ref{eq:alty}, we classify each point into $K$ classes with a SoftMax layer and extract the feature of the last but one layer as the embedding. The learning rate is 1e-3 and the optimizer is Adam~\cite{kingma2014adam}.

\subsection{Pipeline}
The whole pipeline is shown in Figure~\ref{fig:pipeline}. We first infer dense embeddings from human labeled raw annotations. Then fidelity error maps are calculated by summing embedding distances to human labeled keypoints. Non Minimum Suppression is conducted to form a potential set of keypoints. These keypoints are then projected onto 2D subspace with t-SNE and verified by humans.

\section{Tasks and Benchmarks}
In this section, we propose two keypoint prediction tasks: keypoint saliency estimation and keypoint correspondence estimation. Keypoint saliency estimation requires evaluated methods to give a set of potential indistinguishable keypoints while keypoint correspondence estimation asks to localize a fixed number of distinguishable keypoints. 

\subsection{Keypoint Saliency Estimation}


\paragraph{Dataset Preparation}  
For keypoint saliency estimation, we only consider whether a point is the keypoint or not, without giving its semantic label. Our dataset is split into train, validation and test sets with the ratio 70\%, 10\%, 20\%.

\paragraph{Evaluation Metrics} 
Two metrics are adopted to evaluate the performance of keypoint saliency estimation. Firstly, we evaluate their mean Intersection over Unions~\cite{teran20143d} (mIoU), which can be calculated as
\begin{align}
\begin{split}
\label{eq:iou}
    IoU = \frac{TP}{TP + FP + FN}.
\end{split}
\end{align}
mIoU is calculated under different error tolerances from 0 to 0.1. Secondly, for those methods that output keypoint probabilities, we evaluate their mean Average Precisions (mAP) over all categories. 

\paragraph{Benchmark Algorithms}
We benchmark eight state-of-the-art algorithms on point cloud semantic analysis: PointNet~\cite{qi2017pointnet}, PointNet++~\cite{qi2017pointnet++}, RSNet~\cite{huang2018recurrent}, SpiderCNN~\cite{xu2018spidercnn}, PointConv~\cite{wu2019pointconv}, RSCNN~\cite{liu2019relation}, DGCNN~\cite{wang2019dynamic} and GraphCNN~\cite{defferrard2016convolutional}. Three traditional local geometric keypoint detectors are also considered: Harris3D~\cite{sipiran2011harris}, SIFT3D~\cite{rister2017volumetric} and ISS3D~\cite{salti2015learning}. 

\begin{figure}[ht]
    \centering
    \includegraphics[width=0.85\linewidth]{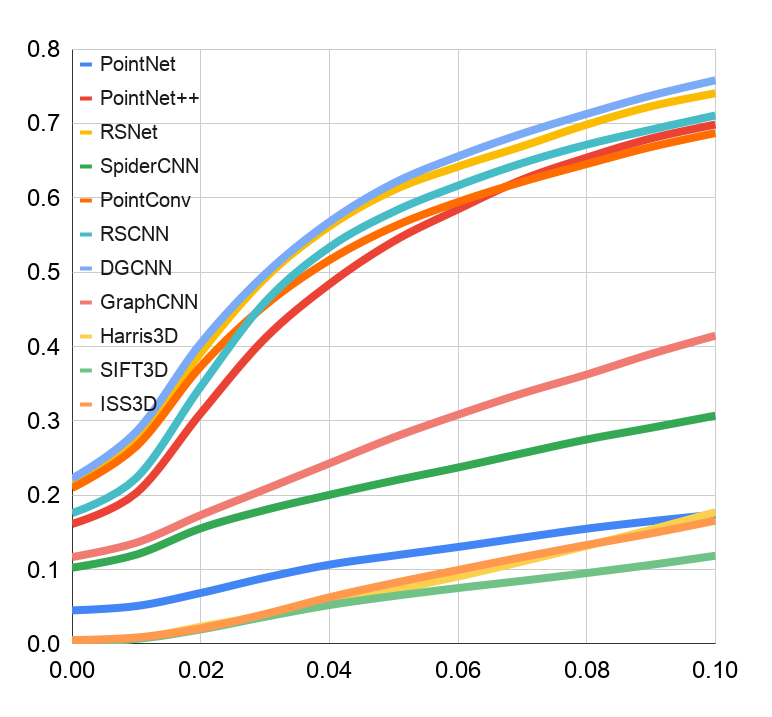}
    \caption{mIoU results under various distance thresholds (0-0.1) for compared algorithms.}
    \label{fig:iou}
\end{figure}

\begin{figure}[ht]
    \centering
    \includegraphics[width=0.85\linewidth]{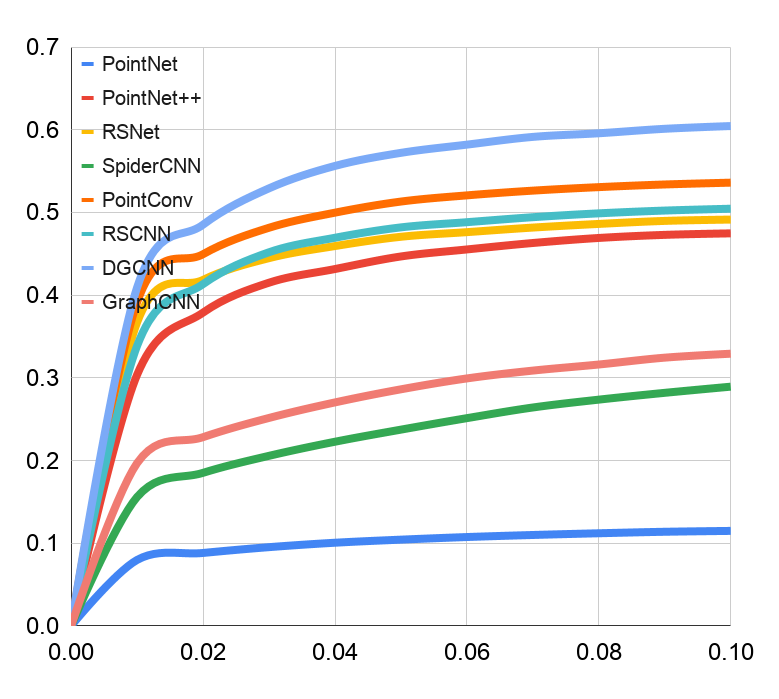}
    \caption{mAP results under various distance thresholds (0-0.1) for compared algorithms.}
    \label{fig:map}
\end{figure}

\paragraph{Evaluation Results}
For deep learning methods, we use the default network architectures and hyperparameters to predict the keypoint probability of each point and mIoU and mAP are adopted to evaluate their performance. For local geometry based methods, mIoU is used. 
Each method is tested with various geodesic error thresholds. In Table~\ref{tab:map}, we report mIoU and mAP results under a restrictive threshold 0.01.
Figure ~\ref{fig:iou} shows the mIoU curves under different distance thresholds from 0 to 0.1 and Figure~\ref{fig:map} shows the mAP results. We can see that under a restrictive distance threshold 0.01, geometric and deep learning methods both fail to predict qualified keypoints.

Figure~\ref{fig:vis_detect} shows some visualizations of the results from RSNet, RSCNN, DGCNN, GraphCNN, ISS3D and Harris3D. Deep learning methods can predict some of ground-truth keypoints while the predicted keypoints are sometimes missing. For local geometry based methods like ISS3D and Harris3D, they give much more interest points spread over the entire model while these points are agnostic of semantic information. Learning discriminative features for better localizing accurate and distinct keypoints across various objects is still a challenging task.

\subsection{Keypoint Correspondence Estimation}
Keypoint correspondence estimation is a more challenging task, where one needs to predict not only the keypoints, but also their semantic labels. The semantic labels should be consistent across different objects in the same category.

\paragraph{Dataset Preparation}  
For keypoint correspondence estimation, each keypoint is labeled with a semantic index. For those keypoints that do not exist on some objects, index -1 is given. Similar to SyncSpecCNN~\cite{yi2017syncspeccnn}, the maximum number of keypoints is fixed and the data split is the same as keypoint saliency estimation.

\begin{figure*}[t]
    \centering
    \includegraphics[width=0.9\linewidth]{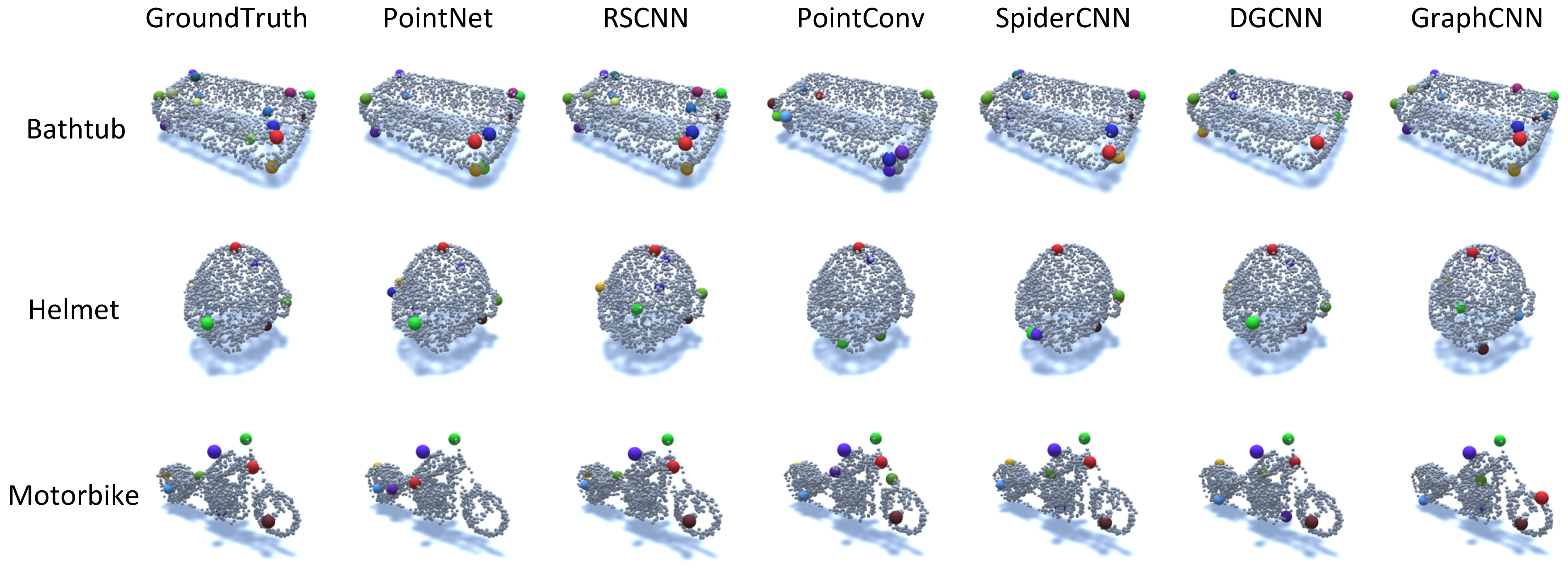}
    \caption{Visualizations of detected keypoints and their semantic labels. Same colors indicate same semantic labels.}
    \label{fig:vis_corr}
\end{figure*}

\begin{table*}[t]
\begin{center}
\resizebox{\textwidth}{!}{
\begin{tabular}{l|c|c|c|c|c|c|c|c|c|c|c|c|c|c|c|c|c}
\hline
~ & \textbf{Airplane}  & \textbf{Bath} & \textbf{Bed} & \textbf{Bottle} & \textbf{Cap} & \textbf{Car} & \textbf{Chair} & \textbf{Guitar} & \textbf{Helmet} & \textbf{Knife} & \textbf{Laptop} & \textbf{Motor} & \textbf{Mug} & \textbf{Skate} & \textbf{Table} & \textbf{Vessel} & \textbf{Average} \\
\hline
\textbf{PointNet} & 65.4 & \textbf{57.2} & 55.4 & \textbf{78.9} & 16.7 & \textbf{66.6} & 43.2 & \textbf{65.4} & \textbf{42.6} & 32.6 & 76.1 & 62.1 & 60.7 & 50.4 & 63.8 & 42.5 & 55.0 \\
\textbf{PointNet++} & 64.1 & 45.4 & 44.2 & 10.2 & 18.8 & 55.1 & 38.0 & 55.4 & 23.6 & 43.8 & 64.6 & 49.9 & 36.4 & 42.3 & 57.1 & 37.9 & 42.9 \\
\textbf{RSNet} & 68.9 & 56.0 & \textbf{69.2} & 78.2 & \textbf{47.9} & 65.7 & 49.4 & 63.7 & 33.8 & \textbf{56.9} & \textbf{80.3} & \textbf{67.4} & \textbf{61.7} & \textbf{69.2} & \textbf{72.3} & 46.7 & \textbf{61.7} \\
\textbf{SpiderCNN} & 54.9 & 36.8 & 43.9 & 50.4 & 0.0 & 47.1 & 36.9 & 37.7 & 5.6 & 32.2 & 63.4 & 36.6 & 16.4 & 15.8 & 61.0 & 36.8 & 36.0 \\
\textbf{PointConv} & \textbf{70.3} & 54.4 & 58.6 & 64.6 & 25.0 & 65.7 & \textbf{51.1} & 62.6 & 16.2 & 47.4 & 74.2 & 61.7 & 51.2 & 51.8 & 69.5 & 46.7 & 54.4 \\
\textbf{RSCNN} & 66.8 & 47.9 & 52.4 & 59.4 & 18.8 & 56.5 & 45.0 & 60.0 & 21.3 & 51.4 & 66.9 & 52.8 & 29.5 & 51.7 & 64.8 & 41.2 & 49.2 \\
\textbf{DGCNN} & 66.8 & 51.5 & 56.3 & 8.9 & 39.6 & 58.6 & 46.0 & 62.4 & 36.1 & 52.1 & 73.9 & 57.7 & 51.4 & 52.3 & 70.5 & \textbf{46.8} & 51.9 \\
\textbf{GraphCNN} & 41.8 & 19.5 & 35.4 & 32.1 & 8.3 & 36.9 & 27.0 & 28.9 & 8.8 & 20.0 & 65.9 & 34.4 & 15.6 & 26.2 & 17.3 & 22.1 & 27.5 \\
\hline
\end{tabular}}
\end{center}   
\caption{PCK results under distance threshold 0.01 for various deep learning networks.}
\label{tab:pck}
\end{table*}

\paragraph{Evaluation Metric}
The prediction of network is evaluated by the percentage of correct keypoints (PCK), which is used to evaluate the accuracy of keypoint prediction in many previous works~\cite{yi2017syncspeccnn, sung2018deep}.

\begin{figure}[h]
    \centering
    \includegraphics[width=0.85\linewidth]{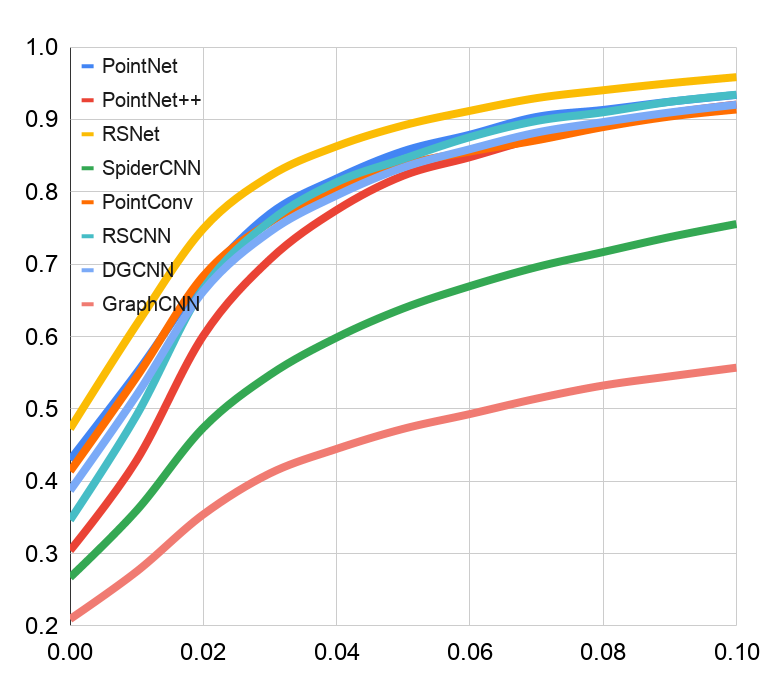}
    \caption{PCK results under various distance thresholds (0-0.1) for compared algorithms.}
    \label{fig:pck}
\end{figure}

\paragraph{Benchmark Algorithms}
We benchmark eight state-of-the-art algorithms: PointNet~\cite{qi2017pointnet}, PointNet++\cite{qi2017pointnet++}, RSNet~\cite{huang2018recurrent}, SpiderCNN~\cite{xu2018spidercnn}, PointConv~\cite{wu2019pointconv}, RSCNN~\cite{liu2019relation}, DGCNN~\cite{wang2019dynamic} and GraphCNN~\cite{defferrard2016convolutional}.

\paragraph{Evaluation Results}
Similarly, we use the default network architectures. Table~\ref{tab:pck} shows the PCK results with error distance threshold 0.01. 
Figure~\ref{fig:pck} illustrates the percentage of correct points curves with distance thresholds varied from 0 to 0.1. RS-Net performs relatively better than the other methods under all thresholds. However, all eight methods face difficulty in giving exact consistent semantic keypoints under a threshold of 0.01.

Figure~\ref{fig:vis_corr} shows some visualizations of the results for different methods. Same colors denote same semantic labels. We can see that most methods can accurately predict some of keypoints. However, there are still some missing keypoints and inaccurate localizations.

Keypoint saliency estimation and keypoint correspondence estimation are both important for object understanding. Keypoint saliency estimation gives a spare representation of object by extracting meaningful points. Keypoint correspondence estimation establishes relations between points on different objects. From the results above, we can see that these two tasks still remain challenging. The reason is that object keypoints from human perspective are not simply geometrically salient points but abstracts semantic meanings of the object.

\section{Conclusion}
In this paper, we propose a large-scale and high-quality KeypointNet dataset. In order to generate ground-truth keypoints from raw human annotations where identification of their modes are non-trivial, we transform the problem into an optimization problem and solve it in an alternating fashion. By optimizing a fidelity loss, ground-truth keypoints, together with their correspondences are generated. In addition, we evaluate and compare several state-of-the-art methods on our proposed dataset and we hope this dataset could boost the semantic understanding of 3D objects.

\clearpage
\section*{Acknowledgements}
This work is supported in part by the National Key R\&D Program of China (No. 2017YFA0700800), National Natural Science Foundation of China under Grants 61772332, 51675342, 51975350, SHEITC (2018-RGZN-02046) and Shanghai Science and Technology Committee.
{\small
\bibliographystyle{ieee_fullname}
\bibliography{egbib}
}

\end{document}